\documentclass[10pt,twocolumn,letterpaper]{article}

\usepackage{wacv}
\usepackage{times}
\usepackage{epsfig}
\usepackage{graphicx}
\usepackage{amsmath}
\usepackage{amssymb}
\usepackage{booktabs}
\usepackage{multirow} 
\usepackage{multicol} 
\usepackage{tabularx}
\usepackage{adjustbox}
\usepackage{bigstrut}
\usepackage[accsupp]{axessibility}  

\def\wacvPaperID{4} 

\wacvapplicationstrack 

\wacvfinalcopy 

\ifwacvfinal
\usepackage[breaklinks=true,bookmarks=false]{hyperref}
\else
\usepackage[pagebackref=true,breaklinks=true,colorlinks,bookmarks=false]{hyperref}
\fi

\begin{document}

\title{Finger-NestNet: Interpretable Fingerphoto Verification on Smartphone using Deep Nested Residual  Network }

\author{Raghavendra Ramachandra 
\quad Hailin Li\\
Norwegian University of Science and Technology (NTNU), Norway\\
{\tt\small E-mail: \{raghavendra.ramachandra, hailin.li\} @ ntnu.no
}}

\maketitle
\thispagestyle{empty}

\begin{abstract}
Fingerphoto images captured using a smartphone are successfully used to verify the individuals that have enabled several applications. This work presents a novel algorithm for fingerphoto verification using a nested residual block - Finger-NestNet. The proposed Finger-NestNet architecture is designed with three consecutive convolution blocks followed by a series of nested residual blocks to achieve reliable fingerphoto verification. This paper also presents the interpretability of the proposed method using four different visualization techniques that can shed light on the critical regions in the fingerphoto biometrics that can contribute to the reliable verification performance of the proposed method. Extensive experiments are performed on the fingerphoto dataset comprised of 196 unique fingers collected from 52 unique data subjects using an iPhone6S. Experimental results indicate the improved verification of the proposed method compared to six different existing methods with EER = 1.15\%.     
\end{abstract}

\section{Introduction}
\begin{figure}[t!]
\begin{center}
\includegraphics[width=1.0\linewidth]{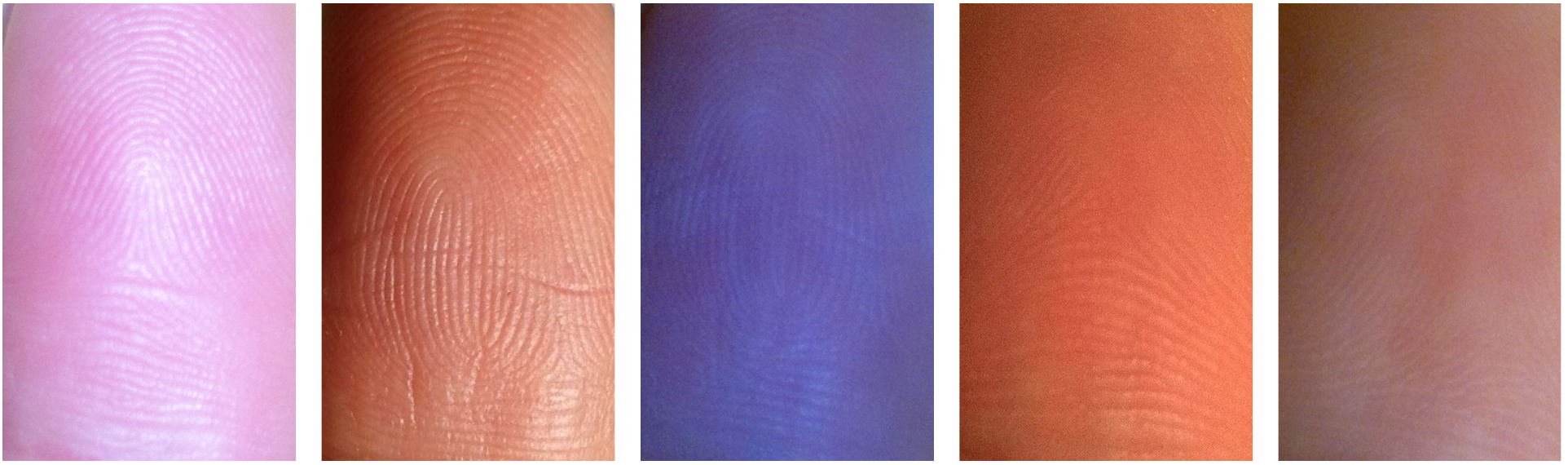}
\end{center}
   \caption{Illustration of fingerphoto captured using smartphone with varying environment conditions(brightness of light)}
\label{fig:IntroFigure}
\end{figure}

The rapid evolution of smartphone technology with highly sophisticated cameras aided reliable biometric authentication. Even though some smartphones in the market have dedicated biometrics sensors, the popularity of smartphone camera-based biometrics has paved the way for several applications because of the scalability, flexibility, and user convenience. This has resulted in several smartphone applications based on the different biometric characteristics such as face \cite{rattani2018survey}, periocular \cite{sharma2022periocular}, visible iris \cite{raja2015smartphone}, multimodal biometrics \cite{ramachandra2019smartphone} and fingerphotos \cite{weissenfeld2022case}.  Among these biometric characteristics, fingerphoto biometrics has gained significant interest as an alternative to touch-based fingerprint authentication due to the COVID-19 pandemic. The fingerphoto authentication employs the built-in camera (normally back-side) to capture the image that can be used for the authentication. Figure \ref{fig:IntroFigure} shows the example of a fingerphoto image captured using an iPhone 6S back camera with various indoor and outdoor scenarios. 

Fingerphoto verification is a challenging task due to unconstrained capture scenarios. The capture quality varies significantly depending on the capture conditions like indoor, outdoor, flash on/off and external light. In addition, the motion blur during fingerphoto capture is highly anticipated due to the small micro-motion of the figure. Further, the fingerphoto images are captured with varied orientation and distance, introducing the challenges of resolving the valley with ridges that further limits the applicability of extracting the minutia-based features. Therefore, the fingerphoto verification systems indicate a degraded performance compared to the conventional fingerphoto verification system.

\begin{table*}[htbp]
  \centering
  \caption{State-of-the-art fingerphoto recognition methods on smartphone}
  \resizebox{\linewidth}{!}{%
    \begin{tabular}{|p{12.355em}|c|p{23.785em}|c|}
    \hline
    Authors & \multicolumn{1}{p{3.215em}|}{year} & Method & \multicolumn{1}{p{14.145em}|}{Database and types of attacks} \bigstrut\\
    \hline
     Lee et al.  \cite{lee2006preprocessing} & 2006  & Color based segmentation and contrast between ridges and valleys &  \multicolumn{1}{p{14.145em}|}{150 fingers, 840 images.}\bigstrut\\
    \hline

    Derawi et.al \cite{derawi2011fingerprint} & 2012  & Matching using VeriFinger SDK &  \multicolumn{1}{p{14.145em}|}{1320  images from each capture device}\bigstrut\\
 
    \hline
     Li et.al \cite{li2012testing} & 2012  & Matching using VeriFinger SDK and NFIS &  \multicolumn{1}{p{14.145em}|}{2100 images from 100 fingers}\bigstrut\\
     \hline
    Raghavendra et. al \cite{raghavendra2013scaling} & 2013  & Mean Shift Segmentation (MSS) and matching through minutiae feature based method &  \multicolumn{1}{p{14.145em}|}{1800 samples from 25 subjects}\bigstrut\\
    \hline
    Tiwari and Gupta \cite{carney2017multi} & 2015  & Matching using scale-invariant features &  \multicolumn{1}{p{14.145em}|}{156 images from 50 fingers}\bigstrut\\
    \hline
    Sankaran et.al \cite{sankaran2015smartphone} & 2015  & ScatNet &  \multicolumn{1}{p{14.145em}|}{5100 images from 128 fingers}\bigstrut\\
    \hline
     Carney et.al \cite{carney2017multi} & 2017  & Four-finger capturing scheme &  \multicolumn{1}{p{14.145em}|}{1400 touchless fingerprints}\bigstrut\\
    \hline
    Chopra et.al \cite{chopra2018unconstrained} & 2018  & ResNet50 representation based matching approach &  \multicolumn{1}{p{14.145em}|}{3450 fingerphotos from 230 fingers}\bigstrut\\
    \hline
    Pankaj et al \cite{wasnik2018baseline}& 2018  & LBP, HOG, BSIF and VeriFinger SDK &  \multicolumn{1}{p{14.145em}|}{720 fingerphotos from 48 fingers}\bigstrut\\
    \hline
    Deb et.al \cite{deb2018matching}& 2019  & Evaluation of Fingerphoto-to-slap-fingerprint matching &  \multicolumn{1}{p{14.145em}|}{309 subjects, 7,976 images}\bigstrut\\

    \hline
   Wild et.al \cite{wild2019comparative} & 2019  & Verifinger SDK &  \multicolumn{1}{p{14.145em}|}{4310 fingerprint images from 108 users}\bigstrut\\
    \hline
    Birajadar et.al \cite{birajadar2019towards} & 2019  & Source-AFIS, NBIS-NIST, Verifinger SDK &  \multicolumn{1}{p{14.145em}|}{200 subjects}\bigstrut\\
   \hline
    Al-Nima et.al \cite{al2020exploiting} & 2020  & Shallow CNN model to recognize a middle finger and index finger &  \multicolumn{1}{p{14.145em}|}{IIITD database}\bigstrut\\
    \hline
   Malhotra et.al \cite{malhotra2020matching} & 2020  & Deep Scattering Network based feature extraction, and Random Decision Forest to authenticate finger-selfies &  \multicolumn{1}{p{14.145em}|}{19456 images (17024 fingerphotos and 2432 live scan fingerprints) from 304 fingers}\bigstrut\\
    \hline 
    Kauba et.al \cite{kauba2021towards} & 2021  & TensorFlow Object Detection API and transfer learning, MobileNetV2 &  \multicolumn{1}{p{14.145em}|}{Training, validation, and test datasets consisted of 1176 images, 168 images, and 784 images, respectively}\bigstrut\\
    \hline
    Attrish et.al \cite{attrish2021contactless} & 2021  & Fuse CNN-based and minutiae based feature score & \multicolumn{1}{p{14.145em}|}{IITI-CFD containing 105 train and 100 test subjects} \bigstrut\\
    \hline
    Priesnitz et.al \cite{priesnitz2022mobile} & 2022  & A fingerprint recognition workflow which can process four fingers and useability issues     & \multicolumn{1}{p{14.145em}|}{1360 samples from 29 subjects} \bigstrut\\
    \hline
    
    \end{tabular}%
    
    }
  \label{tab:SOTA}%
\end{table*}%

The smartphone-based fingerphoto verification is exhaustively studied in the literature, resulting in several algorithms. Table \ref{tab:SOTA} summarizes the fingerphoto verification techniques developed based on smartphone data. Early work on fingerphoto verification is proposed in \cite{lee2006preprocessing} that explored the applicability of fingerphoto for authentication. Since then, several algorithms based on both local features like minutiae and global features like Local Binary Patterns (LBP) \cite{wasnik2018baseline}, Binarized Statistical Image Features (BSIF) \cite{wasnik2018baseline}, Histogram of Gradients (HoG) \cite{wasnik2018baseline}, Scale Invariant features \cite{tiwari2015touch}, Scattered wavelets \cite{malhotra2020matching}. The advances in deep learning have further motivated the researchers to introduce Convolutional Neural Networks (CNN) for fingerphoto verification on a smartphone. To this extent, the transfer learning is performed using MobileNet V2 architecture for fingerphoto verification \cite{kauba2021towards}. The use of a shallow CNN network with single-layer convolution \cite{al2020exploiting} is introduced, which has indicated a reliable verification performance. Further, the hybrid combination of deep features and minutiae-based features \cite{attrish2021contactless} are introduced for reliable fingerphoto verification. Recent work on fingerphoto verification introduced novel features based on the haemoglobin \cite{nakazaki2022fingerphoto} in the finger that can be quantified using color channels. 

\begin{figure*}[htp]
\begin{center}
\includegraphics[width=1.0\linewidth]{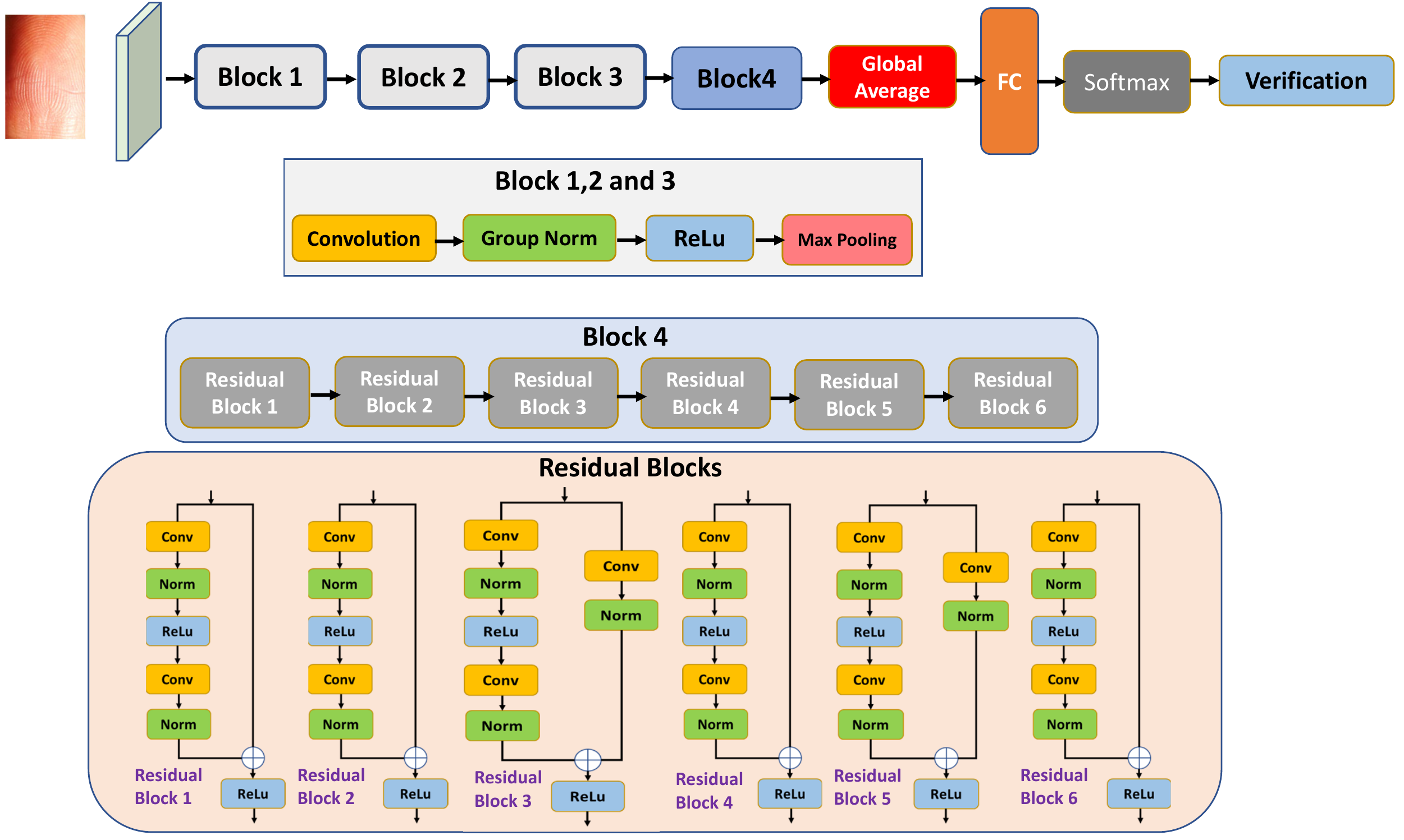}
\end{center}
   \caption{Block diagram of the proposed Finger-NestNet for contactless fingerphoto verification}
\label{fig:Propos}
\end{figure*}

This research proposes a novel algorithm for fingerphoto verification from smartphone-captured data. Based on the available literature, the recent approaches are based on the pre-trained deep CNN; however, the fingerphoto verification still indicates poor performance. Local features like minutiae depend more on the quality of the fingerphoto, and similar observation holds for global features-based fingerphoto verification systems. Further, we also present a detailed analysis of how interpreting and visualizing the activations resulted in the comparison score computed by the proposed method. To the best of our knowledge, this is the first work on the fingerphoto that discusses the interpretability of results and paves the way towards explainability. The following are the main contributions of this work: 

\begin{itemize}
\item	Proposed a novel contactless fingerphoto verification algorithm based on the Finger-NestNet designed using nested residual blocks.
\item	Introduced a new fingerphoto dataset comprised of 196 unique fingers collected from 52 unique data subjects. The fingerphoto dataset is collected using the rear camera of the iPhone 6S smartphone in multiple sessions with an indoor scenario. 
\item	Extensive experiments are carried out by comparing the proposed method with six different existing approaches. The existing approaches include both academic systems based on the deep learning and Commercial-Off-The-Shelves (COTS) system including NeuroTechnology Verifinger \cite{verifinger}. 
\item	Extensive analysis of interpreting the proposed method is presented using four different visualising techniques. 
\end{itemize}

The rest of the paper is organised as follows: Section \ref{sec:pro} presents and discusses the proposed Finger-NestNet architecture, Section \ref{sec:exp} discusses both qualitative and quantitative results of the proposed and state-of-the-art (SOTA) techniques. An extensive analysis of four different visualising techniques is presented to interpret the outcome of the proposed method. Finally, Section \ref{sec:conc} draws the conclusion.    

\section{Proposed Method}
\label{sec:pro}

Figure \ref{fig:Propos} shows the functional block diagram of the proposed Finger-NestNet architecture for contactless fingerphoto verification. The goal of the proposed method is to leverage the residual blocks\cite{he2016deep} to achieve reliable fingerphoto verification. We are motivated to design our network with stacked residual blocks as they can learn the identity function through the skip connection and address the vanishing gradients. Further, the stacked use of residual blocks has proven to guarantee the generalization of unseen data \cite{8984747}.    


As shown in Figure \ref{fig:Propos}, the proposed architecture has three consecutive layers of convolution-group normalization-ReLu-Max Pooling that can learn the high-level features.   The first block is designed to have a convolution layer with a kernel size of $21 \times 21$, which is then reduced by half in the subsequent blocks (e.g., kernel size of $14 \times 14$ in block-2 and of $7 \times 7$ in block 3). The use of dissimilar kernels allows the network to learn different high-level features. We have employed group normalization after each convolution layer to independently normalize the mini-batch of data across each observation that can overcome the sensitivity to network initialization. The residual blocks are connected following the three blocks of the convolutions. The six different residual blocks are connected serially with skip connections. The residual blocks employed in this work have both direct and indirect connections passed through the convolution layer that is summed together at the output. The residual blocks 1, 2, 4 and 5 indicate a similar architecture design but vary with the parameters to allow high-level feature learning. The residual block-1 will connect the skip connection directly from input to output, and the main stem is comprised of convolution-group normalization-ReLu(Rectified Linear units)-convolution-group normalization layers connected in series. The outcome of the additional layer that combines the features from the main stem and the skip connection is further passed through the non-linear activation ReLu that improves the robustness to vanishing gradients. The convolution layer in the main stem in residual block 1 is kernel size $3 \times 3$ with the feature maps size of 32. The residual block-2 is identical to residual block-1. However, with residual block-3, we introduce the convolution-group normalization layer in the skip connection to extract the coarse features further. The convolution layer employed in the skip connection has a kernel size of $1 \times 1$ with a feature map size of 64. The residual block 4 is identical to residual block-2, but the number of feature map sizes is increased to 64. Residual block 5 and 6 shares the same architecture that of residual block 3 and 4, respectively. However, the feature size of a residual block of 5 and 6 correspond to 128. The schematic increase in the feature map size allows the network to precisely capture the robust non-linear feature maps. The outcome of the residual block is then normalized, and a fully connected layer is implemented with 128 nodes. 

\begin{table}[htbp]
  \centering
  \caption{Quantitative performance of the proposed contactless fingerphoto verification}
  \resizebox{.85\linewidth}{!}{%
    \begin{tabular}{|p{15.785em}|c|}
    \hline
    Algorithms & \multicolumn{1}{p{5.855em}|}{EER(\%)} \bigstrut\\
    \hline
    AlexNet Features  \cite{kauba2021towards, attrish2021contactless} & 46.12 \bigstrut\\
    \hline
    DenseNet Features \cite{kauba2021towards, attrish2021contactless} & 24.78 \bigstrut\\
    \hline
    MobileNet Features \cite{kauba2021towards, attrish2021contactless} & 8.26 \bigstrut\\
    \hline
    ResNet50 Features \cite{kauba2021towards, attrish2021contactless} & 15.33 \bigstrut\\
    \hline
    Deep Scattering Network \cite{malhotra2020matching} & 9.33 \bigstrut\\
    \hline
    COTS \cite{verifinger} & 3.84 \bigstrut\\
    \hline
    Proposed Method & 1.15 \bigstrut\\
    \hline
    \end{tabular}%
    }
  \label{tab:Table1}%
\end{table}%
 
Finally, the SoftMax function is used at the output layer corresponding to the number of unique fingerphotos. In this work, we have employed cross-entropy as it is well suited for applications with more than two classes \cite{martinez2018taming}. The proposed Finger-NestNet is optimized using Adam optimizer \cite{kingma2014adam} that takes advantage of adaptive gradient and Root Mean Square (RMS) propagation.

\section{Experiments and Results}
\label{sec:exp}
This section discusses the qualitative and quantitative results of the proposed contactless fingerphoto verification system. The quantitative performance of the proposed method is benchmarked against deep learning-based state-of-the-art techniques that include the deep scattering network  \cite{malhotra2020matching}, and the pre-trained deep CNNs \cite{kauba2021towards, attrish2021contactless}. We compare the performance of the proposed method with the COTS  (NeuroTechnology Verifinger) \cite{verifinger}.
The verification performance of the contactless fingerphoto algorithms is presented in terms of False Non-Match Rate (FNMR), False Match Rate (FMR) and Equal Error Rate (EER). The qualitative results are presented by interpreting the proposed Finger-NestNet using four different visualization methods such as Grad-Cam \cite{selvaraju2017grad}, Occlusion Sensitivity \cite{zeiler2014visualizing}, Locally Interpretable Model-agnostic Explanations (LIME) \cite{ribeiro2016should} and Gradient Attribution \cite{ancona2017towards}. 

Experiments are carried out on the contactless fingerphoto dataset introduced in \cite{wasnik2018baseline} that is further augmented to have 196 unique fingers corresponding to the 52 data subjects. The data is collected using the iPhone 6S with the back camera that can record video at 240 fps with the flash turned on. Session-1 consists of 200 images per unique fingerphoto, and Session-2 consists of 200 images per unique fingerphoto. The fingerphotos are collected in indoor scenarios with standard lighting settings in the two sessions. The contactless fingerphoto systems are trained on the Session-1 data and tested solely on the Session-2 data.

\begin{figure}[htp]
\begin{center}
\includegraphics[width=1.0\linewidth]{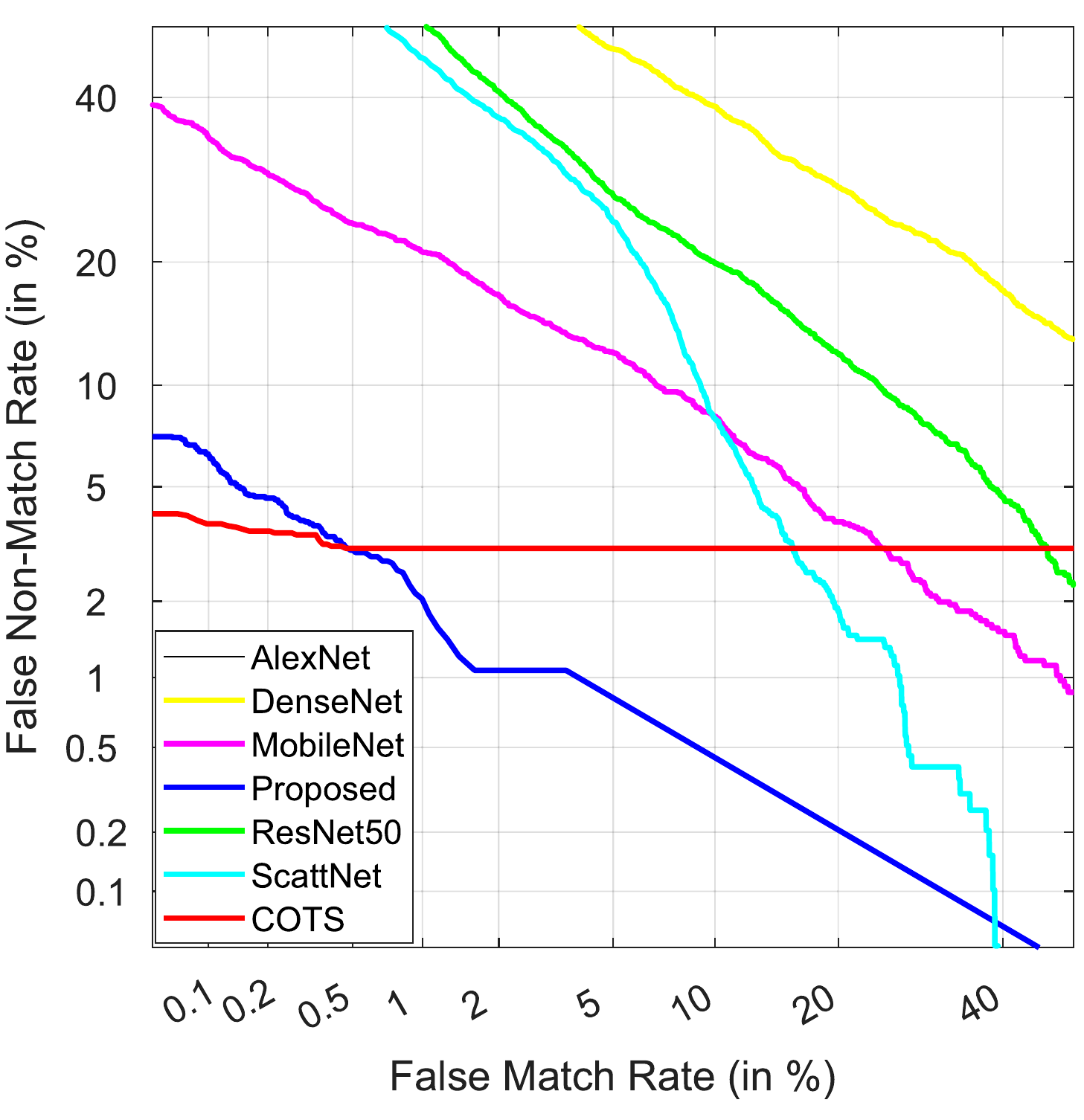}
\end{center}
   \caption{DET performance of the proposed and SOTA contactless fingerphoto method}
\label{fig:ROC}
\end{figure}
\begin{figure*}[htp]
\begin{center}
\includegraphics[width=1.0\linewidth]{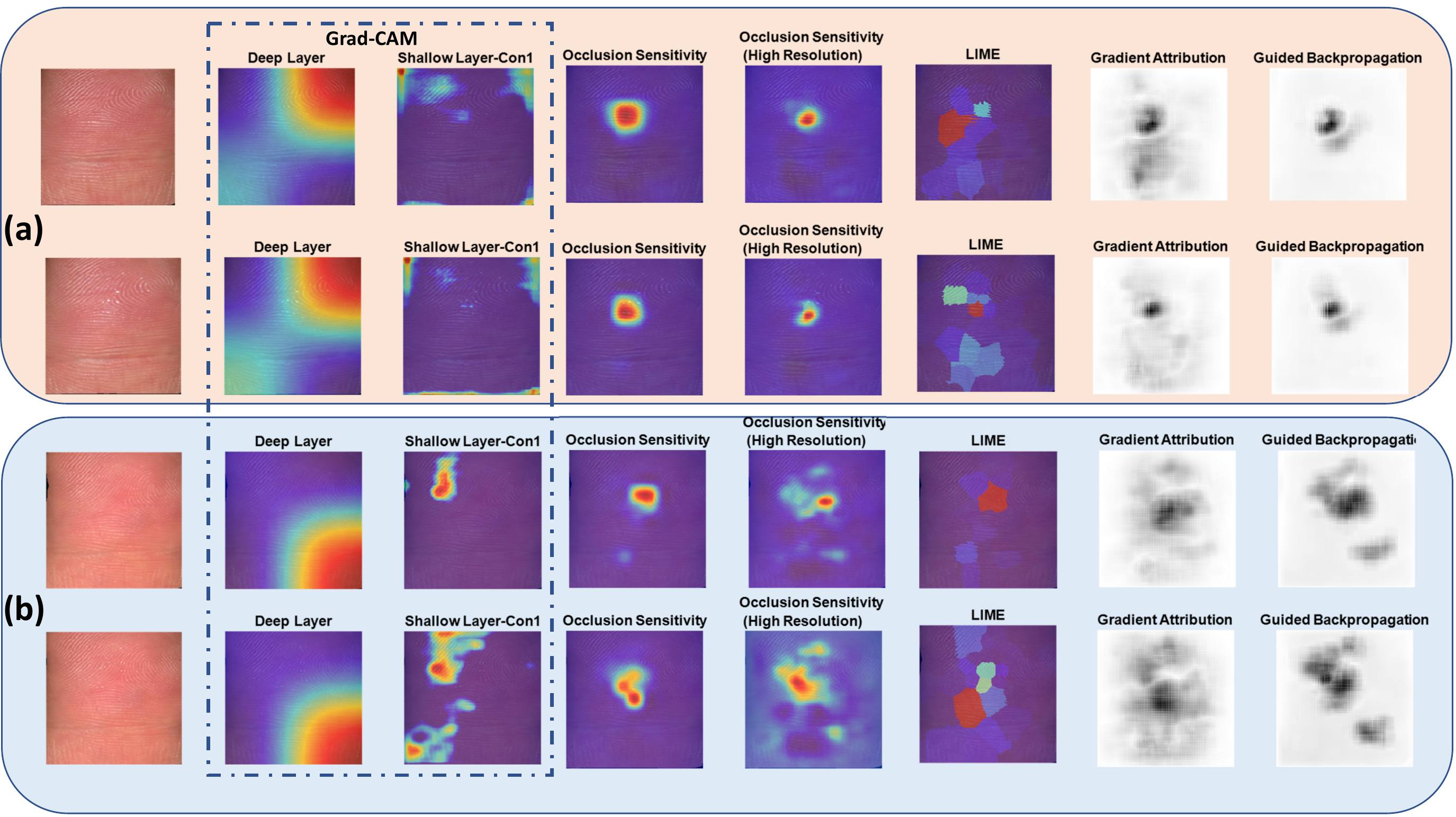}
\end{center}
   \caption{Visualisation of the proposed Finger-NestNet with matched verification (a) high verification match (b) normal verification match}
\label{fig:Interprt3}
\end{figure*}

Table \ref{tab:Table1} indicates the quantitative performance of the proposed method and the state-of-the-art for smartphone-based contactless fingerphoto verification. Figure \ref{fig:ROC} shows the ROC curves indicating the verification performance of the proposed and existing contactless fingerphoto verification. 
The performance of the proposed method is compared against five different deep learning-based approaches that are well exploited in the literature for reliable contactless fingerphoto verification. Based on the obtained results, it can be observed that: 
\begin{itemize}
\item The state-of-the-art deep learning-based methods vary with the type of pre-trained networks employed. 
\item Among the pre-trained deep learning approaches, the AlexNet-based pre-trained network has indicated degraded performance. The best performance among the deep learning approaches is noted with the MobileNet features with EER = 8.25\%.
\item Deep scattering network indicated comparable results than the deep feature-based methods with EER = 9.33\%.
\item The proposed method has indicated the best performance with EER = 1.15\%. Thus, the proposed method performs better than the existing contactless fingerphoto verification approaches. 

\end{itemize}

\begin{figure*}[htp]
\begin{center}
\includegraphics[width=1.0\linewidth]{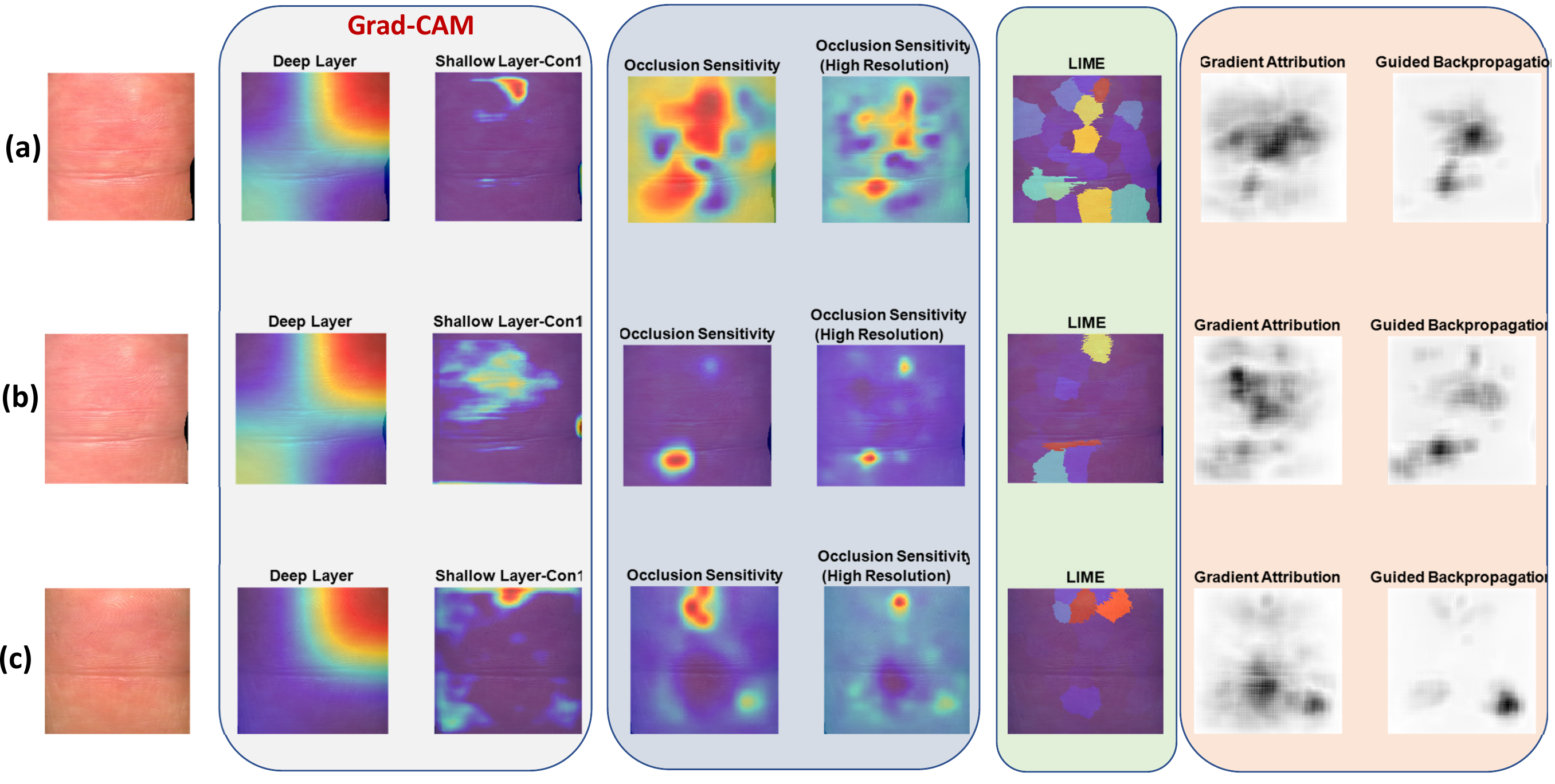}
\end{center}
   \caption{Visualisation of the proposed Finger-NestNet for (a) Enroled sample (b) probe sample (correctly matched to enroled sample) (c) Probe sample (falsely matched to enroled sample)}
\label{fig:example2}
\end{figure*}

To interpret the performance of the proposed Finger-NestNet, we have evaluated four different techniques to understand better how the proposed network can translate the given contactless fingerphoto to the decision. In this work, we have visualized the proposed Finger-NestNet with two different experiments: (a) the genuine comparison that has resulted in high comparison scores and normal comparison scores that has passed the verification threshold (b) match and miss-matched comparison. 

Figure \ref{fig:Interprt3} shows the qualitative results interpreting the decision achieved by the proposed  Finger-NestNet. Figure \ref{fig:Interprt3} (a) visualize the network activations when the two fingerphotos from the same finger are compared had yielded a high comparison score, whereas Figure \ref{fig:Interprt3} (b) shows the visualization of the network when two fingerphotos from the same finger resulted in a normal value of comparison score. As indicated in the Figure \ref{fig:Interprt3} following can be noticed: 
\begin{itemize}

\item The highest match scores are attributed to the activations in the identical region of the fingerphoto image. The Grad-CAM map (see Figure \ref{fig:Interprt3} (a)) indicates the larger gradient values in the top regions of the fingerphotos, especially from the last ReLU layer of the proposed network. Further, the Grad-CAM map of the earlier region (first convolution layer) highlights the edge and line type features at the same finger region. Visualizing occlusion sensitivity between two fingerphoto images highlights the same region of the two fingerphoto images that have contributed to the comparison scores. The high-resolution occlusion sensitivity that will highlight the more precision region also indicates the similarity between the two fingerphotos that are compared. This further justifies the high comparison scores. Further, it is also interesting that high activations are noted around the core region of the fingerphoto samples. Next, we consider the third visualization method using the locally interpretable model-agnostic explanations (LIME) technique. The red area in Figure \ref{fig:Interprt3} (a) indicates the highest important region that can contribute to the decision. It is interesting to note that the LIME method also indicated the similar areas (with different importance) between the fingerphotos that have resulted in the high value of comparison score. The red colour maps can be noted for the same regions (even though an area of the region is different). Lastly, we present the visualization with gradient attribution that can generate a pixel-resolution map corresponding to the most critical pixels in the fingerphoto images. Figure \ref{fig:Interprt3} (a) indicates the gradient activation in the same region that has contributed to the highest comparison score. Thus, based on the four different visualizing methods, we notice that the activations in the same regions have contributed to the highest comparison score. 
\item Figure \ref{fig:Interprt3} (b) illustrates the case for the reduced comparison score magnitude. Here, we can notice that even a small change in the activation region due to the change in image characteristics will reflect a reduction of the comparison score magnitude.
\item It is interesting to notice that the core region in the fingerphoto has contributed mainly to achieving a high comparison score. Further, this observation appears to be consistent with all four visualization methods employed in this work. Thus, the best performance of the proposed method can be attributed to the availability of good-quality information in the fingerphoto core region. 
\end{itemize}

 Figure \ref{fig:example2} shows the illustration of the visualization techniques on both correctly and falsely matched fingerphotos samples by the proposed Finger-NestNet.  Figure \ref{fig:example2} (a) shows the visualization indicating the activations of the proposed method with enrolled fingerphoto, Figure \ref{fig:example2} (b) probe samples that are correctly verified and Figure \ref{fig:example2} (c) probe samples that are falsely verified to the enrolled sample.  We can notice from Figure \ref{fig:example2} (c) that the activations are similar (even though not the same) to that of the enroled fingerphoto image that has resulted in the false verification.

\section{Conclusions}
\label{sec:conc}
Reliable fingerphoto biometric smartphone verification is challenging due to unconstrained data capture. This work presents a novel method for fingerphoto biometric verification using nested residual blocks. The interpretability of the proposed method is exclusively discussed using four different visualizing techniques. The visualization methods consistently indicate the importance of the core region in the fingerphoto image that has contributed to improved performance. Extensive experiments are carried out on the fingerphoto dataset comprised of 196 unique fingers collected from 52 unique data subjects using iPhone 6FS. The performance of the proposed method is compared with six different existing methods including COTS. Experimental results demonstrate the best performance of the proposed method with EER = 1.15\%.  

{\small
\bibliographystyle{ieee_fullname}
\bibliography{egbib}
}

\end{document}